\documentclass[10pt,journal,compsoc]{IEEEtran}

\ifCLASSOPTIONcompsoc
  \usepackage[nocompress]{cite}
\else
  \usepackage{cite}
\fi

\ifCLASSINFOpdf
\else
\fi

\usepackage{amsmath,amssymb,amsfonts}
\usepackage{algorithmic}
\usepackage{array}
\usepackage[pdftex]{graphicx}
\usepackage{textcomp}
\usepackage[table]{xcolor}
\usepackage{multirow}

\def\BibTeX{{\rm B\kern-.05em{\sc i\kern-.025em b}\kern-.08em
    T\kern-.1667em\lower.7ex\hbox{E}\kern-.125emX}}
\usepackage{listings}

\definecolor{codegreen}{rgb}{0,0.6,0}
\definecolor{codegray}{rgb}{0.5,0.5,0.5}
\definecolor{codepurple}{rgb}{0.58,0,0.82}
\definecolor{backcolour}{rgb}{0.95,0.95,0.92}

\lstdefinestyle{mystyle}{
    backgroundcolor=\color{backcolour},   
    commentstyle=\color{codegreen},
    keywordstyle = {\color{magenta}},
    keywordstyle = [2]{\color{lime}},
    keywordstyle = [3]{\color{yellow}},
    keywordstyle = [4]{\color{teal}},
    numberstyle=\tiny\color{codegray},
    stringstyle=\color{codepurple},
    basicstyle=\ttfamily\footnotesize,
    breakatwhitespace=false,         
    breaklines=true,                 
    captionpos=b,                    
    keepspaces=true,                 
    numbers=left,                    
    numbersep=5pt,                  
    showspaces=false,                
    showstringspaces=false,
    showtabs=false,                  
    tabsize=2
}

\begin{document}

\title{ParFormer: A Vision Transformer with Parallel Mixer and Sparse Channel Attention Patch Embedding }
\author{\IEEEauthorblockN{Novendra Setyawan\textsuperscript{1,6}, Ghufron Wahyu Kurniawan\textsuperscript{2}, Chi-Chia Sun\textsuperscript{3, *}, Jun-Wei Hsieh\textsuperscript{4}\\ Jing-Ming Guo\textsuperscript{5}, Wen-Kai Kuo\textsuperscript{1} }
\IEEEauthorblockA{\textit{\textsuperscript{1}Department of Electro-Optics, National Formosa University, Taiwan}\\
\textit{\textsuperscript{2}Department of Electrical Engineering, National Formosa University, Taiwan} \\
\textit{\textsuperscript{3}Department of Electrical Engineering, National Taipei University, Taiwan} \\
\textit{\textsuperscript{4}College of Artificial Intelligence and Green Energy, National Yang Ming Chiao Tung University, Taiwan} \\
\textit{\textsuperscript{5}Department of Electrical Engineering, National Taiwan University of Science and Technology, Taiwan} \\
\textit{\textsuperscript{6}Department of Electrical Engineering, University of Muhammadiyah Malang, Indonesia} \\
\textit{chichiasun@gm.ntpu.edu.tw\textsuperscript{*}}
} 
}

\markboth{IEEE Transactions on Cognitive and Developmental System (Pre-Print)}%
{Setyawan \etal{}: ParFormer: A Vision Transformer with Parallel Mixer and Sparse Channel Attention Patch Embedding}

\maketitle

\begin{abstract}
In recent years, both Convolutional Neural Networks (CNNs) and Transformers have achieved remarkable success in computer vision tasks. However, their deep architectures often lead to high computational redundancy, making them less suitable for resource-constrained environments, such as edge devices. This paper introduces ParFormer, a novel vision transformer that addresses this challenge by incorporating a Parallel Mixer and a Sparse Channel Attention Patch Embedding (SCAPE). By combining convolutional and attention mechanisms, ParFormer improves feature extraction. This makes spatial feature extraction more efficient and cuts down on unnecessary computation. The SCAPE module further reduces computational redundancy while preserving essential feature information during down-sampling. Experimental results on the ImageNet-1K dataset show that ParFormer-T achieves 78.9\% Top-1 accuracy with a high throughput on a GPU that outperforms other small models such as MobileViT-S, FasterNet-T2, and EdgeNeXt-S. Specifically, ParFormer-T achieves 2.56$\times$ higher throughput than MobileViT-S, 0.24\% faster than FasterNet-T2, and 1.79$\times$  higher than EdgeNeXt-S. For edge device deployment, ParFormer-T excels with a throughput of 278.1 images/sec, which is 1.38 $\times$ higher than EdgeNeXt-S and 2.36$\times$ higher than MobileViT-S, making it highly suitable for real-time applications in resource-constrained settings. The larger variant, ParFormer-L, reaches 83.5\% Top-1 accuracy, offering a balanced trade-off between accuracy and efficiency, surpassing many state-of-the-art models. In COCO object detection, ParFormer-M achieves 40.7 AP for object detection and 37.6 AP for instance segmentation, surpassing models like ResNet-50, PVT-S and PoolFormer-S24  with significantly higher efficiency. These results validate ParFormer as a highly efficient and scalable model for both high-performance and resource-constrained scenarios, making it an ideal solution for edge-based AI applications.
\end{abstract}
\section{Introduction}
\label{sec:intro}

In recent years, deep learning models, particularly Convolution Neural Network (CNN) and Transformers, have garnered significant interest and demonstrated remarkable achievement in computer vision and natural language processing \cite{carion2020end, dai2021up, arnab2021vivit}. There has been a clear observation in the domain of deep learning models that as the network grows deeper, it becomes possible to acquire more sophisticated features to enhance the performance. In classical CNN, ResNet50 \cite{he2016deep} and VGG-16 \cite{simonyan2014very} have around 25 million and 138 million parameters, respectively, assigned to image feature extraction and classification tasks. Moreover, the vanilla Vision Transformer (ViT) \cite{dosovitskiy2020image} needs 85 million to 632 million parameters to do the similar ImageNet classification task. Their forward propagation requires a significant amount of computation, which is slow inference and not easily accessible in many particular fields of application. Application of CNN models in low-cost settings, such as mobile and edge devices, is highly difficult because of the restricted memory, processing power, and battery life. Therefore, our work is to construct a lightweight and efficient deep learning model that minimizes the amount of computation while nonetheless ensuring fast inference and high performance.

Numerous studies \cite{chen2023run, zhang2020split, yun2024shvit, han2020ghostnet, wang2023ldcnet} contend that feature redundancy may be the root of CNN or Transformer models' inefficiency. For instance, the intermediate layer of ResNet50 exhibits significant redundancy in feature maps across the channels. It indicates that elevated parameters in ResNet50 do not enhance the model's accuracy or generalization capability. Not only in CNN, different approaches, such as transformer models, may lead to redundancy. The tokenization of image patches and subsequent attention mechanisms such as multi-head self attention (MHSA) in DeiT \cite{touvron2021training} and Swin Transformers \cite{liu2021swin} also suffer feature redundancy across the head. Each patch in a transformer may represent visual patterns that are too similar to neighboring patches, causing the MHSA to focus on repetitive information rather than diverse. In \cite{yun2024shvit} shows that there are 64.8\% feature similarities when using 6 heads and get lower when the number of heads is reduced. It shows inefficiency using many heads in MHSA that used much memory to token folding or reshaping during MHSA. Therefore, there are several ongoing research projects on network design to deal with it.

Several studies have been done to eliminate computational redundancy by designing efficient modules in networks. In Fasternet \cite{chen2023run} proposed partial convolution (PConv) to process a quarter feature channel only. Using $3 \times 3$ conventional convolution, process spatial information along with a quarter channel followed by FFN as a channel mixer to construct a Fasternet block. Along with that, InceptionNeXt \cite{yu2024inceptionnext} uses 3 depth-wise convolutions (DWConv) with different kernel sizes as a spatial mixer to process the $3/8$ channel. A set of $3 \times 3, 1 \times 11$, and $11 \times 1$ DWConv was proposed to make a high-throughput network and a high-receptive field to deal with segmentation tasks. In \cite{yun2024shvit} proposed single head attention (SHA) to learn spatial information for a quarter feature channel. The purpose of SHA in SHViT to keep the other channel untouched is to avoid feature redundancy such as in MHSA. 

\begin{figure*}[!ht]
    \centering
    \includegraphics[width=1\linewidth]{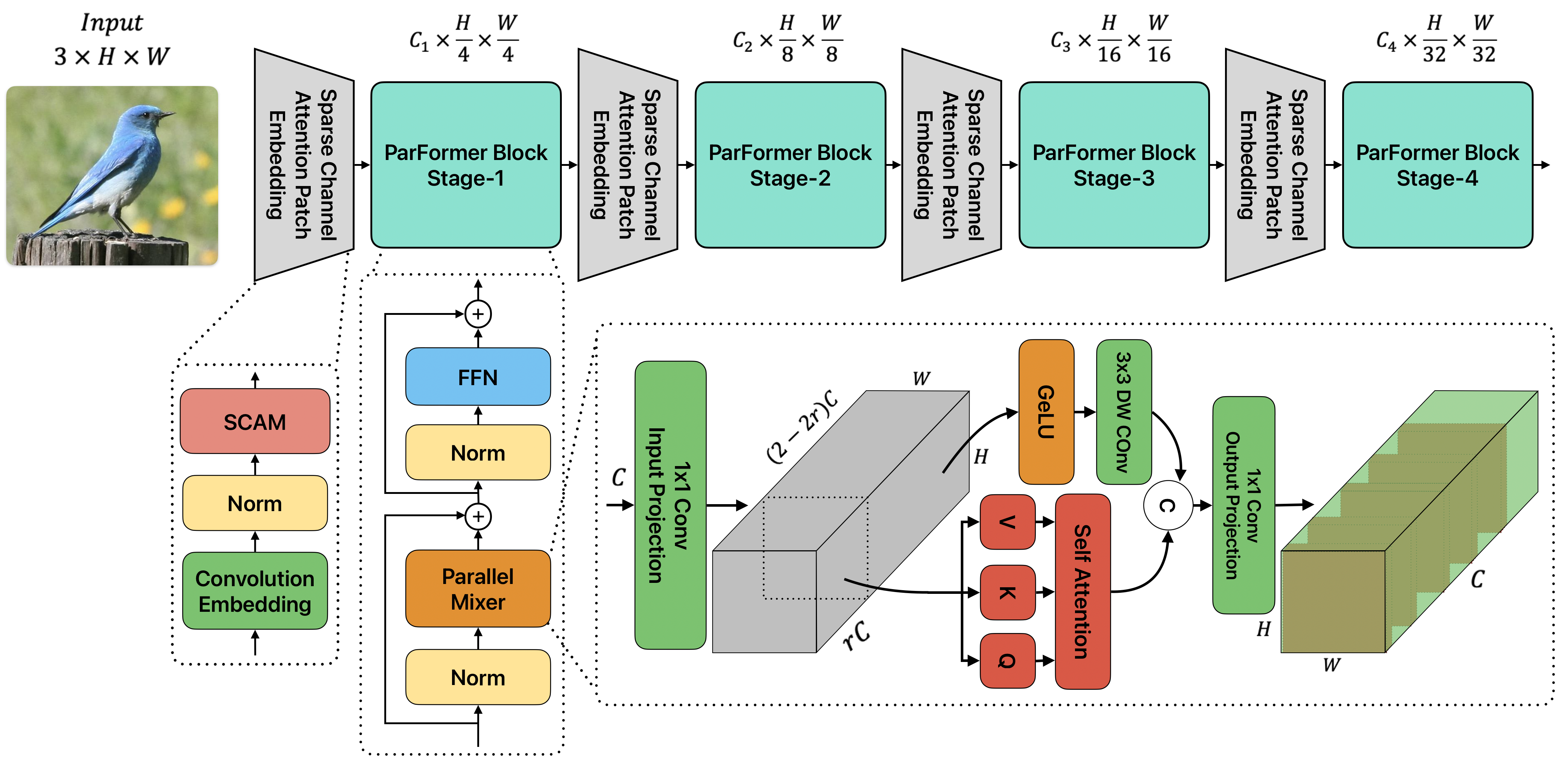}
    \caption{\textbf{The overall ParFormer Architecture}, ParFormer used pyramid architecture with 4 stages. The feature spatial dimension is decreased from stride 4 to stride 32 using SCAPE module with increased channel dimension in each stage are $[C_1, C_2, C_3, C_4]$. In each stage, the Parallel Mixer is used with different ratio configurations.}
    \label{fig:parformer_arch}
\end{figure*}

The presence of abundant and redundant information in the feature maps of deep neural networks frequently ensures a thorough comprehension of the input data. Redundancy in feature maps may be a crucial attribute for an effective deep neural network. Even though several works try to avoid feature redundancy by processing several channels only and keeping others untouched, they need a large number of channels to keep the accuracy maintained, which leads to slower inference and high parameter. Rather than avoiding unnecessary feature maps, we prefer to incorporate them in a cost-effective manner by proposing a ParFormer as shown in Figure \ref{fig:parformer_arch}. To reduce feature and computational redundancy, we proposed a novel Parallel Mixer and Sparse Channel Attention Patch Embedding (SCAPE). The contributions of ParFormer are summarized as follows:
\begin{itemize}
    \item A parallel mixer is proposed by combining two different spatial mixer to generate multi-receptive field features. Combining self-attention, which can capture long-range sequence information, with DWConv, the local feature extractor, to construct a parallel mixer. Then project both features to ensure two different pieces of information will propagate across the channel.
    \item Integrating the Sparse Channel Attention Module (SCAM) into the Patch Embedding Module to create the down-sampling module known as Sparse Channel Attention Patch Embedding (SCAPE). The SCAPE module is engineered to prevent information loss during down-sampling across the channel. Implementing linear operations in SCAM transforms the interconnections between channels from a dense configuration to a sparse one, thereby conserving parameters and computational resources. 
    \item Based on MetaFormer architecture with the Parallel Mixer and SCAPE module, we design a novel basic architectural block ParFormer Encoder. We also design a network structure called ParFormer, which is constructed by stacking multiple ParFormer Encoder.
\end{itemize}

The subsequent sections of this article are structured as follows. Section II presents an overview of pertinent research on efficient network architectures for CNNs and Transformers. In Section III, we delineate the proposed network architectures, which encompass the SCAPE module alongside SCAM, Parallel Mixer, and ParFormer. Section IV presents numerous comparative experiments to validate the efficacy of the proposed network model. Finally, conclusions and expectations are drawn in Section V.


\par 

\section{Related Works}
\subsection{Vision Transformers} The integration of transformers from natural language processing (NLP) into the field of computer vision was first proposed by \cite{dosovitskiy2020image}. Since then, this idea has garnered considerable attention and has been used to various activities within the domain. Numerous transformer topology have been proposed \cite{ wang2021pyramid, wang2022pvt, liu2021swin, liu2022swin} with the aim of enhancing precision and effectiveness in diverse applications. Generally, these techniques prioritize the attention mechanism of the transformer architecture due to its computational complexity and lack of continuity being identified as the primary challenge. For example, the Pyramid Vision Transformer (PVT) \cite{wang2021pyramid} introduced a modification known as Spatial Reduction Attention (SRA), which replaces the Vanilla Attention mechanism. SRA achieves a reduction in computational complexity by employing key and value reductions, while still preserving the global receptive field characteristic of the ViT. On the other hand, the transformer model exhibits a deficiency in terms of local continuity. To address this issue, alternative local approaches, such as the Swin Transformer \cite{liu2021swin}, adopt a strategy of using non-overlapping windows. These windows are gradually shifted to increase the receptive field, thereby capturing interactions across several phases. Therefore, the self-attention mechanism has constraints in its ability to efficiently capture information that spans across great distances.

\subsection{Parallel Mixer} In numerous works, the attention-based model transformer is combined with a convolution network or another type of attention model in order to reduce computational complexity and address the issue of discontinuity \cite{mehta2021mobilevit, li2022efficientformer, vasu2023fastvit, maaz2022edgenext, li2023rethinking}. In a general context, the process of combining or hybridising can be categorised into two distinct groups: serial combination and parallel combination. Within the framework of a sequential methodology, the integration of the attention module and convolution can yield a serial design \cite{mehta2021mobilevit, maaz2022edgenext, vasu2023fastvit}. In the initial phase of its architecture, the convolution network performs local field extraction, but in the later stage, it employs the attention module to obtain global receptive capabilities. The purpose of this approach is to ensure that the architecture remains lightweight while still being able to retrieve and store both local and global information. On the other hand, this approach involves implementing parallelism between the local and global extractors within the attention module. The PLG-ViT \cite{ebert2023plg} model exhibits parallelism in its attention mechanisms, specifically in the W-MSA of Swin Transformer as a local extractor and the SRA of PVT as a global extractor. Another approach in Lite Transformer \cite{wu2020lite}, combine vanilla attention as a long range feature capturing with convolution network as local feature capturing in transformer encoder. Next approach \cite{wu2023parallel}, using combination transformer encoder block from ViT \cite{dosovitskiy2020image} and inverted bottleneck block from MobilenetV2 \cite{sandler2018mobilenetv2} as a fusion block in Human Pose Estimation and Running Movement Recognition. In \cite{jain2020attention}, use transformer block and dilated convolution block and combined using fusion score as and ensemble network for sketch recognition using vector images. However, most of them combine two spatial mixer separately without any mechanism to allow propagation across the channel between them.

\subsection{Patch Embedding} In the initial implementation of the transformer, the image is minimally processed by linear projection, as opposed to the CNN-based model which utilizes convolution with stride in the block to reduce the spatial size of the image. In the upcoming iteration of Transformer, the vision transformer architecture, which is employed in PVT\cite{wang2021pyramid} and CVT\cite{wu2021cvt}, incorporates convolution patch embedding . In order to enhance the local coherence, the technique of overlap patch embedding is employed in PVTv2 \cite{wang2022pvt} and CVT\cite{wu2021cvt}. Subsequently, it has been demonstrated that the "patchify" technique may effectively improve the performance of the CNN architecture  in ConvNeXt\cite{liu2022convnet}. This enhancement enables the ConvNeXt design to attain comparable results to the Swin Transformer\cite{liu2021swin}. In our methodology, we draw inspiration from the Squeeze-and-Excitation technique \cite{hu2018squeeze} to incorporate sparse channel attention into the patch embedding process. This is done with the aim of improving the learnability in image patching and reducing loss of information while downsampling. 

\begin{figure*}[!ht]
    \centering
    \includegraphics[width=16cm]{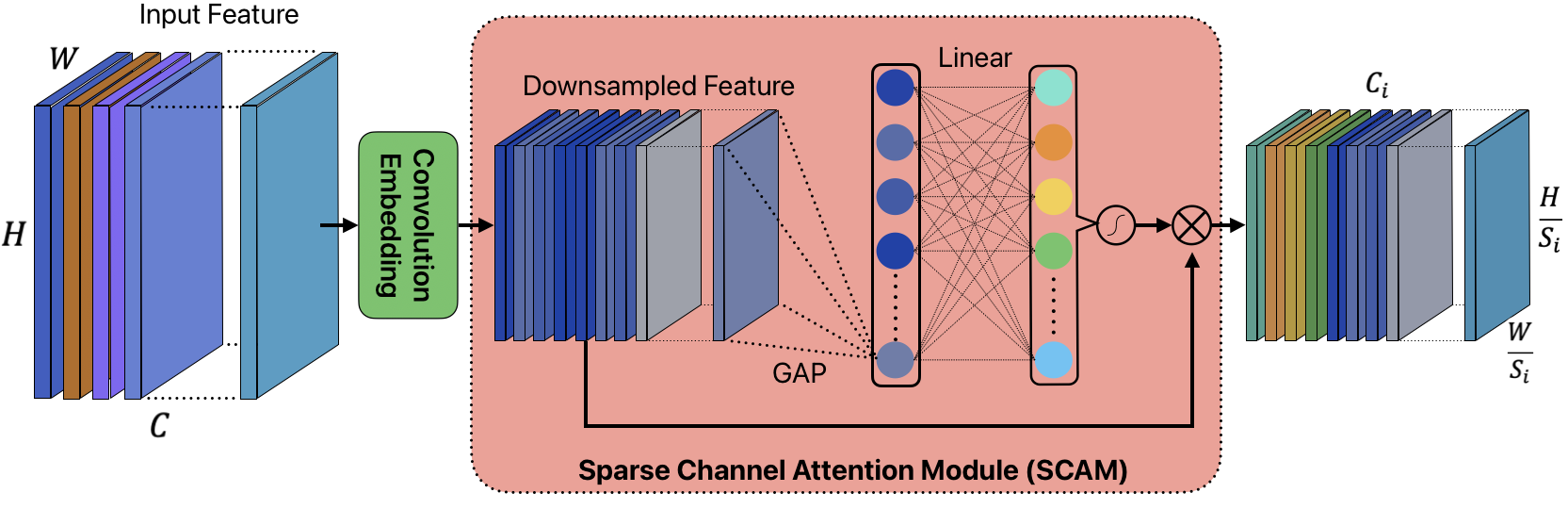}
    \caption{\textbf{Sparse Channel Attention Patch Embedding.} The input feature can be the image or feature maps from the previous stage that downsized using stride convolution. Non-linearity function after linear layer in the SCAM module is Sigmoid function. Finally the element wise product is used to re-calibrated the downsized feature maps.}
    \label{fig:CAPE}
\end{figure*}
\section{Proposed Method}

The main idea of a novel ParFormer architecture for an efficient vision model is reducing the redundancy in image features. The ParFormer model incorporates the SCAPE module and parallel mixer technique, which could generate low-redundancy feature maps. The Sparse Channel Attention Patch Embedding (SCAPE) is constructed by integrating the sparse channel attention module into the patch embedding to reduce feature information loss while downsampling. The parallel mixer, combination of attention and convolution mixer, can extract the feature map information efficiently with different receptive field.

\subsection{Sparse Channel Attention Patch Embedding (SCAPE)} 
Transformer-based vision models \cite{liu2021swin, wu2021cvt, wang2022pvt} often use patch embedding as a layer for downsampling space and increasing the number of channels. In the original Vision Transformer (ViT)\cite{dosovitskiy2020image}, to transform a 2D image into a 1D token sequence, patch embedding utilized convolution and folded the 2D feature into a 1D sequence. In the recent lightweights ViT, an overlap convolution layer is used, keeping the feature in 2D shape since the feature folding process or tensor reshaping consumes more memory. Along with that, ParFormer architecture used convolution to process the given input image or the feature map from the previous stage $X_{i}$ with the size $H\times W\times C_{i}$. 
\begin{equation}
    X_{i+1}=SCAM(PatchEmbedding(X_{i})).
    \label{eq:cape}
\end{equation}
Using the patch embedding in equation \ref{eq:cape}, the input $X_{i}$ will be patched into $X_{i+1}\in \mathbb{R}^{\frac{H}{S_i}\times\frac{W}{S_i}\times C_{i+1}}$ with convolution stride $S_i$, kernel size $2S_i-1$, and padding size $S_i-1$ for the overlapped spatial downsizing and $C_{i+1}$ for the number of feature in the stage-i. The $4 \times 4$ and $2 \times 2$ kernel size are chosen for patch size.

Upon patch embedding, we integrate the SCAM to minimize channel redundancy and information loss caused by the patching process. The Sparse Channel Attention Module, or SCAM, is derived from the classical Squeeze-and-Excitation module\cite{hu2018squeeze}, which can be integrated into the existing model. It has the ability to re-calibrate the feature map in each channel to maintain the interdependence between the channels of feature maps to enhance their representation ability. 

\begin{equation}
    SCAM(X_i)= \sigma(W_s(F_{gap}(X_i))).
    \label{eq:SCAM}
\end{equation}
As depicted in Figure \ref{fig:CAPE}, The SCAM obtains the channel saliency through global average pooling (GAP) to squeeze space dependency. Then, instead of using two fully connected (FC) layers, a single linear operator ($W_s$) is applied to make it efficient. Finally, a nonlinear sigmoid function ($\sigma$) could scale the input features to highlight the useful channels as defined in \ref{eq:SCAM}.

\begin{table*}[ht]
\begin{center}
\caption{All Variant ParFormer Model configurations. \#Blocks denotes number of parformer encoder blocks. Patch size also denotes as SCAPE convolution kernel and stride size}
\begin{tabular}{cccccccc}
\hline
\multicolumn{1}{c|}{\multirow{2}{*}{Stage}} & \multicolumn{1}{c|}{\multirow{2}{*}{Token Size}} & \multicolumn{2}{c|}{\multirow{2}{*}{Layer Specification}} & \multicolumn{3}{c}{ParFormer}  \\ 
\cline{5-8} 
\multicolumn{1}{l|}{}  & \multicolumn{1}{l|}{} & \multicolumn{2}{l|}{} & \multicolumn{1}{c|}{T} & \multicolumn{1}{c|}{S} & \multicolumn{1}{c|}{M} & \multicolumn{1}{c}{L} \\ 
\hline
\multicolumn{1}{l|}{\multirow{6}{*}{Stage-1}} & \multicolumn{1}{l|}{\multirow{6}{*}{$\frac{H}{4}\times\frac{W}{4}$}} & \multicolumn{1}{c|}{\multirow{2}{*}{SCAPE}}   & \multicolumn{1}{l|}{\begin{tabular}[c]{@{}l@{}}Patch Size\end{tabular}} & \multicolumn{4}{c}{7x7, Stride 4} \\ 
\cline{4-8} 
\multicolumn{1}{l|}{} & \multicolumn{1}{l|}{} & \multicolumn{1}{c|}{} & \multicolumn{1}{l|}{Dim ($C$)}  & \multicolumn{1}{c|}{48} & \multicolumn{1}{c|}{64}  & \multicolumn{1}{c|}{96} & \multicolumn{1}{c}{112}  \\ 
\cline{3-8} 
\multicolumn{1}{l|}{} & \multicolumn{1}{l|}{} & \multicolumn{1}{c|}{\multirow{4}{*}{\begin{tabular}[c]{@{}c@{}}ParFormer\\ Encoder\end{tabular}}} & \multicolumn{1}{l|}{ratio $r$} & \multicolumn{1}{c|}{0}  & \multicolumn{1}{c|}{0} & \multicolumn{1}{c|}{0} & \multicolumn{1}{c}{0} \\ 
\cline{4-8} 
\multicolumn{1}{l|}{} & \multicolumn{1}{l|}{} & \multicolumn{1}{c|}{} & \multicolumn{1}{l|}{DWConv Dim ($C_c$)}  & \multicolumn{1}{c|}{ 96 } & \multicolumn{1}{c|}{ 128 } & \multicolumn{1}{c|}{192} & \multicolumn{1}{c}{224}   \\
\cline{4-8}
\multicolumn{1}{l|}{} & \multicolumn{1}{l|}{} & \multicolumn{1}{c|}{} & \multicolumn{1}{l|}{FFN ratio ($\alpha$)}  & \multicolumn{1}{c|}{2} & \multicolumn{1}{c|}{2} & \multicolumn{1}{c|}{2} & \multicolumn{1}{c}{2}   \\ 
\cline{4-8} 
\multicolumn{1}{l|}{} & \multicolumn{1}{l|}{} & \multicolumn{1}{c|}{} & \multicolumn{1}{l|}{\#Blocks} & \multicolumn{1}{c|}{1}  & \multicolumn{1}{c|}{1}  & \multicolumn{1}{c|}{1} & \multicolumn{1}{c}{2} \\ 
\hline
\multicolumn{1}{l|}{\multirow{6}{*}{Stage-2}} & \multicolumn{1}{l|}{\multirow{6}{*}{$\frac{H}{8}\times\frac{W}{8}$}} & \multicolumn{1}{c|}{\multirow{2}{*}{SCAPE}}  & \multicolumn{1}{l|}{\begin{tabular}[c]{@{}l@{}}Patch Size\end{tabular}} & \multicolumn{4}{c}{3x3, Stride 2} \\
\cline{4-8} 
\multicolumn{1}{l|}{} & \multicolumn{1}{l|}{} & \multicolumn{1}{c|}{} & \multicolumn{1}{l|}{Dim ($C$)}  & \multicolumn{1}{c|}{96}  & \multicolumn{1}{c|}{128}  & \multicolumn{1}{c|}{192}  & \multicolumn{1}{c}{224} \\ 
\cline{3-8} 
\multicolumn{1}{l|}{}  & \multicolumn{1}{l|}{}  & \multicolumn{1}{c|}{\multirow{4}{*}{\begin{tabular}[c]{@{}c@{}}ParFormer\\ Encoder\end{tabular}}} & \multicolumn{1}{l|}{ratio $r$} & \multicolumn{1}{c|}{0} & \multicolumn{1}{c|}{0} & \multicolumn{1}{c|}{0} & \multicolumn{1}{c}{0} \\ 
\cline{4-8} 
\multicolumn{1}{l|}{} & \multicolumn{1}{l|}{} & \multicolumn{1}{c|}{} & \multicolumn{1}{l|}{DWConv Dim ($C_c$)} & \multicolumn{1}{c|}{192} & \multicolumn{1}{c|}{256} & \multicolumn{1}{c|}{384} & \multicolumn{1}{c}{448}  \\ 
\cline{4-8}
\multicolumn{1}{l|}{} & \multicolumn{1}{l|}{} & \multicolumn{1}{c|}{} & \multicolumn{1}{l|}{FFN ratio ($\alpha$)}  & \multicolumn{1}{c|}{2} & \multicolumn{1}{c|}{2} & \multicolumn{1}{c|}{2} & \multicolumn{1}{c}{2}   \\ 
\cline{4-8} 
\multicolumn{1}{l|}{} & \multicolumn{1}{l|}{} & \multicolumn{1}{c|}{} & \multicolumn{1}{l|}{\#Blocks} & \multicolumn{1}{c|}{2} & \multicolumn{1}{c|}{2} & \multicolumn{1}{c|}{2} & \multicolumn{1}{c}{4} \\ 
\hline
\multicolumn{1}{l|}{\multirow{6}{*}{Stage-3}} & \multicolumn{1}{l|}{\multirow{6}{*}{$\frac{H}{16}\times\frac{W}{16}$}} & \multicolumn{1}{c|}{\multirow{2}{*}{SCAPE}}  & \multicolumn{1}{l|}{\begin{tabular}[c]{@{}l@{}}Patch Size\end{tabular}} & \multicolumn{4}{c}{3x3, Stride 2} \\ 
\cline{4-8} 
\multicolumn{1}{l|}{} & \multicolumn{1}{l|}{} & \multicolumn{1}{l|}{}  & \multicolumn{1}{l|}{Dim ($C$)} & \multicolumn{1}{c|}{192}  & \multicolumn{1}{c|}{256} & \multicolumn{1}{c|}{384} & \multicolumn{1}{c}{448}\\ 
\cline{3-8}  
\multicolumn{1}{l|}{} & \multicolumn{1}{l|}{} & \multicolumn{1}{l|}{\multirow{6}{*}{\begin{tabular}[c]{@{}c@{}}ParFormer\\ Encoder\end{tabular}}} & \multicolumn{1}{l|}{ratio $r$}  & \multicolumn{1}{c|}{0} & \multicolumn{1}{c|}{$1/4$}  & \multicolumn{1}{c|}{$1/4$} & \multicolumn{1}{c}{$1/4$} \\
\cline{4-8}
\multicolumn{1}{l|}{}  & \multicolumn{1}{l|}{}  & \multicolumn{1}{l|}{} & \multicolumn{1}{l|}{DWConv Dim ($C_c$)} & \multicolumn{1}{c|}{384}  & \multicolumn{1}{c|}{384} &  \multicolumn{1}{c|}{576} & \multicolumn{1}{c}{672}  \\ 
\cline{4-8}
\multicolumn{1}{l|}{}  & \multicolumn{1}{l|}{}  & \multicolumn{1}{l|}{} & \multicolumn{1}{l|}{Attn Dim ($C_a$)} & \multicolumn{1}{c|}{0}  & \multicolumn{1}{c|}{64} &  \multicolumn{1}{c|}{96} & \multicolumn{1}{c}{112}  \\ 
\cline{4-8}
\multicolumn{1}{l|}{}  & \multicolumn{1}{l|}{}  & \multicolumn{1}{l|}{} & \multicolumn{1}{l|}{QK Dim ($C_q, C_k$)} & \multicolumn{1}{c|}{0}  & \multicolumn{1}{c|}{32, 32} &  \multicolumn{1}{c|}{32, 32} & \multicolumn{1}{c}{32, 32}  \\ 
\cline{4-8}
\multicolumn{1}{l|}{} & \multicolumn{1}{l|}{} & \multicolumn{1}{c|}{} & \multicolumn{1}{l|}{FFN ratio ($\alpha$)}  & \multicolumn{1}{c|}{2} & \multicolumn{1}{c|}{2} & \multicolumn{1}{c|}{2} & \multicolumn{1}{c}{2}   \\ 
\cline{4-8} 
\multicolumn{1}{l|}{} & \multicolumn{1}{l|}{} & \multicolumn{1}{l|}{} & \multicolumn{1}{l|}{\#Blocks}  & \multicolumn{1}{c|}{7} & \multicolumn{1}{c|}{7} & \multicolumn{1}{c|}{7} & \multicolumn{1}{c}{9} \\ 
\hline
\multicolumn{1}{l|}{\multirow{6}{*}{Stage-4}} & \multicolumn{1}{l|}{\multirow{6}{*}{$\frac{H}{32}\times\frac{W}{32}$}} & \multicolumn{1}{c|}{\multirow{2}{*}{SCAPE}}  & \multicolumn{1}{l|}{\begin{tabular}[c]{@{}l@{}}Patch Size\end{tabular}} & \multicolumn{4}{c}{3x3, Stride 2} \\ 
\cline{4-8} 
\multicolumn{1}{l|}{} & \multicolumn{1}{l|}{} & \multicolumn{1}{l|}{}  & \multicolumn{1}{l|}{Dim ($C$)} & \multicolumn{1}{c|}{384}  & \multicolumn{1}{c|}{512} & \multicolumn{1}{c|}{768} & \multicolumn{1}{c}{896}\\ 
\cline{3-8}  
\multicolumn{1}{l|}{} & \multicolumn{1}{l|}{} & \multicolumn{1}{l|}{\multirow{6}{*}{\begin{tabular}[c]{@{}c@{}}ParFormer\\ Encoder\end{tabular}}} & \multicolumn{1}{l|}{ratio $r$}  & \multicolumn{1}{c|}{$1/4$} & \multicolumn{1}{c|}{$1/4$}  & \multicolumn{1}{c|}{$1/4$} & \multicolumn{1}{c}{$1/4$} \\
\cline{4-8}
\multicolumn{1}{l|}{}  & \multicolumn{1}{l|}{}  & \multicolumn{1}{l|}{} & \multicolumn{1}{l|}{DWConv Dim ($C_c$)} & \multicolumn{1}{c|}{288}  & \multicolumn{1}{c|}{384} &  \multicolumn{1}{c|}{576} & \multicolumn{1}{c}{672}  \\ 
\cline{4-8}
\multicolumn{1}{l|}{}  & \multicolumn{1}{l|}{}  & \multicolumn{1}{l|}{} & \multicolumn{1}{l|}{Attn Dim ($C_a$)} & \multicolumn{1}{c|}{96}  & \multicolumn{1}{c|}{128} &  \multicolumn{1}{c|}{192} & \multicolumn{1}{c}{224}  \\ 
\cline{4-8}
\multicolumn{1}{l|}{}  & \multicolumn{1}{l|}{}  & \multicolumn{1}{l|}{} & \multicolumn{1}{l|}{QK Dim ($C_q, C_k$)} & \multicolumn{1}{c|}{32, 32}  & \multicolumn{1}{c|}{32, 32} &  \multicolumn{1}{c|}{32, 32} & \multicolumn{1}{c}{32, 32}  \\ 
\cline{4-8}
\multicolumn{1}{l|}{} & \multicolumn{1}{l|}{} & \multicolumn{1}{c|}{} & \multicolumn{1}{l|}{FFN ratio ($\alpha$)}  & \multicolumn{1}{c|}{2} & \multicolumn{1}{c|}{2} & \multicolumn{1}{c|}{2} & \multicolumn{1}{c}{2}   \\ 
\cline{4-8} 
\multicolumn{1}{l|}{} & \multicolumn{1}{l|}{} & \multicolumn{1}{l|}{} & \multicolumn{1}{l|}{\#Blocks}  & \multicolumn{1}{c|}{2} & \multicolumn{1}{c|}{2} & \multicolumn{1}{c|}{2} & \multicolumn{1}{c}{3} \\ 
\hline
\multicolumn{4}{c|}{Classifier Head} & \multicolumn{4}{c}{Global Pool, FC (1280 Hidden) }\\ \hline
\end{tabular}
\label{tab:arch_variant}
\end{center}
\end{table*}

\subsection{Parallel Mixer} The information of the patched image will be extracted into the ParFormer block with repeated sequence and two residual sub-blocks: an Feed Forward Network (FFN) block and a parallel mixer block. The parallel mixer comprises two different spatial operations that enable effective extraction of token information. We proposed combination of single-head attention (SHA) and depth-wise convolution (dwconv) to construct a parallel mixer block. Supposed that $X_i$ is the feature map from SCAPE block, the parallel mixer will projected $A\in \mathbb{R}^{H\times W\times rC_{i}}$ and $V_{dw}\in \mathbb{R}^{H\times W\times (2-2r)C_{i}}$ using convolution operation with $W_{ip}$ is the $1 \times 1$ convolution weight and $r\in[0.0, 1.0]$ is the attention channel number ratio. Unlike \cite{wu2023parallel, jain2020attention} that using two different spatial mixer separately, we proposed convolution projection to ensure efficient propagation of the attention features with dwconv feature as detailed in equation \ref{eq:par_mixer}.
\begin{align}
    ParallelMixer(X_i) &= W_{ip}*Cat(A,V_{dw})
    \label{eq:par_mixer}
\end{align} 
The global attention score $A$ and local convolution feature map $V_{dw}$ processed previously from the following equation
\begin{align}
    A &= V_a\cdot Softmax(Q^T\cdot K) \\
    V_{dw} &= DWConv(\sigma(V_c))
    \label{eq:att_dw}
\end{align} 
where $Q\in \mathbb{R}^{H\times W\times C_{q}}$, $K\in \mathbb{R}^{H\times W\times C_{k}}$, $V_a\in \mathbb{R}^{H\times W\times C_{a}}$, and $V_{c}\in \mathbb{R}^{H\times W\times C_{c}}$ are query, key, attention value, and convolution value. The query and key channel dimensions $C_q$ and $C_k$ are set to 32 as a the maximum following vanilla attention \cite{yu2023metaformer, dosovitskiy2020image}. The channel number of attention value and convolution value $C_a$ and $C_c$ following the channel ratio $r$ with $rC$ for attention channel ratio and $2(C-C_a)$ for dwconv channel as illustrated in Figure \ref{fig:parformer_arch}. The optimal channel number ratio between attention and convolution is set $r=1/4$, as detailed in the ablation study. To expand the input feature channel dimension $X_i$ into parallel mixer, $1 \times 1$ convolution with $W_{ip}$ weight is used as input projection.  The projected input will be divided with the splitting array that corresponds to the channel as described in equation \ref{eq:in_proj}.
\begin{align}
    Q, K, V_a, V_c &= Split(W_{ip}*X_i), 
    \label{eq:in_proj}
\end{align} 

Finally, following \cite{yu2022metaformer}, the residual block with $\lambda_i$ as the layer scale parameter applied after parallel mixer block and followed by residual FFN block as detailed in equation bellow
\begin{align}
    X'_i &= X_i+\lambda_i \odot ParallelMixer(X_i)\\
    X''_i &= X'_i+\lambda_i \odot FFN(X'_i)
    \label{eq:encoder}
\end{align} 
where $X_i', X_i'' \in \mathbb{R}^{H\times W \times C_i} $. This paper's FFN block is similar to the original MLP block used in the Vision Transformer (ViT) with point-wise convolution as linear operation replacement to minimize tensor reshaping \cite{dosovitskiy2020image}. It encompasses a singular activation function, which may be mathematically represented by the subsequent equation.
\begin{equation}
    FFN(X'_i)=(\sigma(Norm(X'_i)*W_{fc1}))*W_{fc2},
\end{equation}
 The $W_{fc1} \in \mathbb{R}^{C_i\times \alpha C_i}$ and $W_{fc2} \in \mathbb{R}^{\alpha C_i\times C_i}$ are learn-able weight with $\alpha$ expansion ratio. The $\sigma$ denotes the activation function that can be configured as $GELU(.)$\cite{hendrycks2016gaussian}, $ReLU(.)$\cite{agarap2018deep}, or another non-linear activation function (\emph{i.e} $SiLU(.)$ \cite{elfwing2018sigmoid}).

 \subsection{Overall Architechture}
In the preceding section, we explain the novel parallel mixer in the ParFormer approach.  This enabled us to assess the efficacy of the model by employing two different token mixers at each stage of the encoder. The ParFormer design as a general-purpose vision transformer utilized four-stage pyramid architecture. The SCAPE is used to downsampling the feature with spatial reduction rates {4, 8, 16, 32} with embedding layer using $7 \times 7$ patch size for the first embedding layer or stem layer and the other embedding or merging layer using $3 \times 3$. For each stage contain a stack of ParFormer encoder that employ a parallel mixer with a different ratio $r$ in every stage and ParFormer variant. For example, parallel mixer with ratio of $r=0$ means all feature channels will be learned from DWConv mixer. If parallel mixer with ratio of $r=1/4$ means 1/4 feature channel will learned from attention based mixer and others in DWConv mixer. We observe that in the early stage will consume higher memory access since then we employ parallel mixer ratio $r=0$ to use DWConv mixer as less complexity then attention for all variant.

Alongside the aforementioned layer, normalization and activation layers are essential for optimal neural network performance. In contrast to the conventional transformer model that use Layer Normalization \cite{lei2016layer}, we utilize batch normalization (BN) to reduce tensor reshaping or folding. Furthermore The advantage of Batch Normalization is its ability to be integrated with neighboring Convolution layers, resulting in expedited inference without compromising efficacy. We empirically select GELU \cite{hendrycks2016gaussian} for the activation layers in all variations of ParFormer, taking into account both execution time and efficacy. The final layers, namely global average pooling, a fully connected layer with 1280 number of hidden, are employed collectively for feature modification and classification.

The overall architecture variant of the ParFormer is shown in Table \ref{tab:arch_variant}. There are four ParFormer variants: tiny (T), small (S), medium (M), and large(L). The blocks in each stage and the feature dimension is manually chosen to format a several models in different size, whose the computational complexity ranges from 0.82G(T) to 6.2G(L). The number of block ratio is following the recipe from ConvNeXt\cite{liu2022convnet} and MetaFormer\cite{yu2023metaformer}.
\begin{table*}[!ht]
\centering
\caption{Comparison of All ParFormer Variant with State-Of-The-Art on ImageNet-1K Dataset. Edge-ONNX is the throughput on NVIDIA Jetson Orin Nano Edge Device using ONNX format.}
\begin{tabular}{ m{3.6cm}|>{\centering}m{1.1cm}|>{\centering}m{1.0cm}|>{\centering}m{1.1cm}|>{\centering}m{0.8cm}|>{\centering}m{0.8cm}|>{\centering}m{0.8cm}|c }
\hline
\multirow{2}{*}{Model}    & \multirow{2}{*}{Img Size}  & \multirow{2}{*}{Params} & \multirow{2}{*}{GFLOPs} & \multicolumn{3}{c|}{Throughput (Img/s)}   & \multirow{2}{*}{Top-1} \\ \cline{5-7} 
                                         &         &               &           &  GPU  & CPU & Edge-ONNX &       \\ \hline
PoolFormer-S12\cite{yu2022metaformer}   & 224 & 11.8 & 1.82 & 2704 & 62.8 & 151.8 & 77.2   \\
PVTv2-B1\cite{wang2022pvt}              & 224 & 14.0 & 2.04 & 1530 & 60.7 & 79.8 & 78.1  \\
MobileViTV2-1.0 \cite{mehta2021mobilevit} & 256 & 4.9 & 1.88 & 1858 & 33.4 & 104.3 & 78.1 \\ 
MobileViT-S \cite{mehta2021mobilevit}   & 256 & 5.6 & 1.89 & 1393  & 32.3 & 82.5 & 78.4  \\
ConvNeXtV2-F\cite{liu2022convnet}       & 224 & 5.2 & 0.78  & 3048 & 97.7 & 115.4 & 78.5  \\
ResNet50\cite{he2016deep}               & 224 & 25.6 & 4.14 & 1680 & 31.3 & 137.4 & 78.8  \\
Fasternet-T2\cite{chen2023run}          & 224 & 14.9 & 1.90 & 3989 & 82.8 & 277.2 & 78.9  \\
\rowcolor{gray!30}
\textbf{ParFormer-T}                    & 224  & 7.4 & 0.82 & 4961 & 106.6 & 278.1 & 78.9 \\ \hline 
EdgeNeXt-S \cite{maaz2022edgenext}      & 256 & 5.6 & 1.26 & 1774 & 80.5 & 116.7  & 79.4 \\
EfficientFormerV2-S1\cite{li2023rethinking} & 224 & 6.2 & 0.68 & 1055 & 63.5 & 185.3 & 79.7 \\
PoolFormer-S24\cite{yu2022metaformer}   & 224 & 21.4 & 3.4 & 1403 & 32.7 & 79.9 & 80.3   \\
ConvNeXtV2-P\cite{liu2022convnet}       & 224 & 9.1 & 1.37 & 2324 & 63.2 & 88.6 & 80.3  \\
MobileViTV2-1.5 \cite{mehta2021mobilevit} & 256 & 10.6 & 4.15 & 1116 & 20.9 & 63.2 & 80.4 \\
FastViT-SA12\cite{vasu2023fastvit}    & 256 & 11.5 & 1.93 & 2125 & 47.0 & 105 & 80.6 \\
\rowcolor{gray!30}
\textbf{ParFormer-S}                   & 224  & 11.9 & 1.48 & 3350 & 78.6 & 193.7 & 80.8 \\ \hline
RepViT-M1.5 \cite{wang2024repvit}      & 224 & 14.0 & 2.27 & 2096 & 40.7 & 119.6  & 81.2 \\
Fasternet-S\cite{chen2023run}          & 224 & 31.2 & 4.55 & 2059 & 40.3 & 157.6 & 81.3 \\ 
Swin-T\cite{liu2021swin}               & 224 & 28.3 & 4.51 & 886 & 26.4 & OOM & 81.3 \\
PoolFormer-S36\cite{yu2022metaformer}  & 224 & 30.9 & 5.00 & 948 & 22.3 & 53.4 & 81.4 \\
ConvNeXtV2-N\cite{liu2022convnet}      & 224 & 15.6 & 2.45 & 1596 & 45.8 & 62.1 & 81.9 \\
EfficientFormerV2-S2\cite{li2023rethinking}& 224 & 12.6 & 1.30 & 569 & 41.9 & 114.5 & 82.0 \\
ConvNeXt-T\cite{liu2022convnet}        & 224 & 28.6 & 4.47 & 1148 & 39.4 & 70.48 & 82.1 \\
PVTv2-B2\cite{wang2022pvt}             & 224 & 25.4 & 3.90 & 866 & 32.7 & 45.5 & 82.1 \\
PoolFormer-M36\cite{yu2022metaformer}  & 224 & 56.17 & 8.80 & 634 & 14.6 & 35.7 & 82.1 \\
InceptionNeXt-T\cite{yu2024inceptionnext}& 224 & 28.1 & 4.20 & 1528 & 32.2 & 94.9 & 82.3 \\
MobileViTV2-2.0\cite{mehta2021mobilevit}& 256 & 18.5 & 7.29 & 786 & 14.3 & 30.9 & 82.4 \\
EfficientFormer-L3\cite{li2022efficientformer}& 224 & 31.3 & 3.93 & 1198 & 28.1 & 96.3 & 82.4 \\
RepViT-M2.3 \cite{wang2024repvit}       & 224 & 22.9 & 4.52 & 1161 & 23.6 & 70.3 & 82.5 \\
FastViT-SA24 \cite{vasu2023fastvit}     & 256 & 21.5 & 3.75 & 1154 & 25.1 & 58.5 & 82.6 \\
\rowcolor{gray!30}
\textbf{ParFormer-M}                    & 224 & 24.2 & 3.17 & 2001 & 42.2 & 117.0  & 82.4 \\ \hline
Swin-S\cite{liu2021swin}                & 224 & 49.6 & 8.77 & 531 & 15.3 & OOM   & 83.0  \\ 
ConvNeXt-B\cite{liu2022convnet}        & 224 & 88.6 & 15.38 & 444 & 14.6 & 29.3 & 83.8 \\
ConvNeXtV2-T\cite{liu2022convnet}      & 224 & 28.6 & 4.47 & 867 & 26.5 & OOM & 83.0 \\
ConvFormer-S18\cite{yu2023metaformer}   & 224 & 26.7 & 3.95 & 1118 & 32.0 & 55.3 & 83.0 \\
Fasternet-L\cite{chen2023run}           & 224 & 93.4 & 15.50 & 657 & 14.0 & 51.3 & 83.5  \\
InceptionNeXt-S\cite{yu2024inceptionnext}& 224 & 49.4 & 8.36 & 811 & 18.0 & 55.4 & 83.5 \\
CAFormer-S18\cite{yu2023metaformer}     & 224 & 26.3 & 3.89 & 963 & 29.8 & 58.8 & 83.6 \\
FastViT-SA36\cite{vasu2023fastvit}      & 256 & 31.5 & 5.56 & 751 & 17.0 & 40.2 & 83.6 \\
\rowcolor{gray!30}
\textbf{ParFormer-L}                    & 224 & 42.3 & 6.2 & 1064 & 21.5 &  61.9 & 83.5 \\ \hline

\end{tabular}
\label{tab:imagenet}
\end{table*}

\section{Experimental Result}
In order to assess the effectiveness of ParFormer, two thorough tests are conducted. The tests involve ImageNet-1K datasets \cite{russakovsky2015imagenet} for classification and COCO Datasets \cite{lin2014microsoft} for object detection and instance segmentation. 

\subsection{Image Classification}
\subsubsection{Preparation} Image classification experiments are conducted using the ImageNet-1K dataset, consisting of 1.28 million training images and 50,000 validation images across 1,000 categories. Our study used a training recipe similar to those of DeiT\cite{touvron2021training} and Swin Transformer\cite{liu2021swin}. The models undergo training for a total of 300 epochs, with a resolution of 224$\times$224. Data augmentation and regularization approaches encompass several methods such as RandAugment \cite{cubuk2020randaugment}, Mixup \cite{zhang2017mixup}, CutMix \cite{yun2019cutmix}, Random Erasing \cite{zhong2020random}, weight decay, Label Smoothing \cite{huang2017densely}, and Stochastic Depth \cite{szegedy2016rethinking}. The hyper-parameters of DeiT \cite{touvron2021training} are primarily adhered to in our approach. We employ the AdamW optimizer \cite{you2019large} with 256 batch size on 3$\times$A6000 GPUs for most ParFormer models. The performance evaluation was conducted without any pre-training on larger datasets like ImageNet-21K or knowledge distillation.

To evaluate the throughput, we select three representative processors that encompass a broad spectrum of computational capabilities: GPU (RTX-3090), CPU (Intel i5-13500), and Edge Device utilizing Jetson Orin Nano (NVIDIA Ampere 1024 CUDA and 32 tensor cores). We present their throughput for inputs with a batch size of 256 for both GPU and CPU. For Edge Device utilizing a batch size of 64 with ONNX Runtime. During inference, the batch normalization layers of the ParFormer model are integrated with their adjacent layers when appropriate.

\subsubsection{ImageNet-1K Classification Result} The performance comparison of various models on the ImageNet-1K dataset, as detailed in Table \ref{tab:imagenet}, underscores several important findings related to throughput, model complexity, and accuracy, especially for edge deployment on the NVIDIA Jetson Orin Nano using the ONNX format. To show the effectiveness of ParFormer, the results are grouped based on accuracy. Our proposed method, ParFormer, demonstrates exceptional balance across these metrics, making it a strong candidate for edge-based AI tasks. When evaluating throughput on GPUs, ParFormer-T stands out with an impressive 4961 images per second, outperforming other CNN or Transformer based models such as ConvNeXtV2-F \cite{woo2023convnext}, Fasternet-T2 
 \cite{chen2023run} and MobileViT series \cite{mehta2021mobilevit, mehta2022separable} in terms of GPU computation efficiency. This indicates that ParFormer is well-suited for real-time image processing on GPU-equipped devices. Even in CPU scenarios, where throughput tends to be significantly lower, ParFormer-T maintains competitive performance with 106.6 img/s, which positions it as a versatile model for environments where GPU resources may be limited.

For edge deployment, the Edge-ONNX throughput is particularly critical, and here, ParFormer-T excels with 278.1 img/s, making it the highest-performing model in this category. This highlights the efficiency of our method in real-world edge applications, where computational resources are constrained. Similarly, ParFormer-S achieves a notable 193.7 img/s in Edge-ONNX throughput compare to other lightweight model such as MobileViTV2-1.5 \cite{mehta2022separable} and FastViT-SA12 \cite{vasu2023fastvit}, further reinforcing the versatility and scalability of our proposed architecture across various edge scenarios.

In terms of model complexity, ParFormer maintains a relatively low computational cost. For instance, ParFormer-T has just 7.4 million parameters and 0.82 GFLOPs, offering a lightweight solution that balances performance and efficiency. Even ParFormer-M and ParFormer-L, which have larger parameter counts (24.2M and 42.3M, respectively), still perform exceptionally well in edge settings while offering strong accuracy, with ParFormer-L achieving 83.5\% top-1 accuracy. While models like CAFormer-S18 \cite{yu2023metaformer}(83.5\%) and Swin-L \cite{liu2021swin} (83.6\%) slightly edge out in accuracy, these models require significantly more parameters and computational resources. In contrast, ParFormer offers highly competitive accuracy while maintaining lower computational complexity, particularly in resource-limited edge scenarios. Lastly, unlike some larger models such as Swin-L \cite{liu2021swin} and ConvNeXtV2-T \cite{woo2023convnext}, which face out-of-memory (OOM) issues during edge deployment, ParFormer models exhibit no such limitations, confirming their suitability for memory-constrained environments.

The results in Table \ref{tab:imagenet} clearly show that ParFormer is an excellent solution for an edge-based deep learning vision model, offering an optimal balance between throughput, accuracy, and computational efficiency. Our proposed ParFormer architecture not only excels in high-throughput scenarios but also addresses the practical constraints of edge deployment, making it a highly viable option for real-world applications.

\subsection{Ablation Study}
In this part, we examine the unique architectural decisions. Table \ref{tab:ablation} presents an ablation study on ParFormer-S, analyzing the effects of key design elements, specifically the Parallel Mixer (PM) ratio and the SCAPE module, which incorporates the SCAM. SCAM is crucial for minimizing information loss across feature channels during downsampling. The study investigates how varying these components affects the model's parameters, computational complexity (GFLOPs), throughput (in images per second on a GPU), and Top-1 accuracy.

The baseline configuration of ParFormer-S includes 11.9 million parameters, 1.47 GFLOPs, a throughput of 3362 images per second, and achieves a Top-1 accuracy of 80.9\%. This serves as the reference point for comparing the effects of modifications in both the Parallel Mixer and SCAM designs.

\begin{table}[!ht]
\centering
\caption{Ablation of ParFormer on ImageNet-1K with 224$\times$224 image size. PM ratio denotes Parallel Mixer ratio with tuple in 4 stage. PE denotes Patch Embedding. GPU denotes the throughput in image/s}
\begin{tabular}{ m{1.4cm}|m{2.2cm}|m{0.7cm}|m{0.8cm}|m{0.6cm}|c } \hline
Ablation                                & Variant & Params        & GFLOPs & GPU & Top-1  \\ \hline
ParFormer-S        & Baseline   & 11.9 & 1.47 & 3362 & 80.9  \\ \hline
\multirow{3}{*}{PM ratio $r$} & [0, 0, 0, 0] & 12.0 & 1.45 & 3435 & 80.4 \\ \cline{2-5}
                       & [0, 0, 1/4, 1/4] & 11.9 & 1.48 & 3362 & 80.9 \\ \cline{2-5}
                       & [0, 0, 1/2, 1/2] & 11.3 & 1.41 & 3428 & 80.2  \\ \hline
\multirow{3}{*}{SCAPE} & SCAM Before PE & 11.6 & 1.48 & 3368 & 80.7 \\ \cline{2-5}
                       & SCAM After PE & 11.9 & 1.48 & 3362 & 80.9 \\ \cline{2-5}
                       & PE wo/ SCAM & 11.5 & 1.48 & 3390 & 80.4 \\ \hline

\end{tabular}
\label{tab:ablation}
\end{table}

\begin{table*}[!ht]
\centering
\caption{Object detection and instance segmentation on COCO val2017 with Mask R-CNN. $AP^b$ and $AP^m$ denote bounding box average precision and mask average precisionm, respectively. The FLOPs (G) are measured at resolution 1280 $\times$ 800.
}
\begin{tabular}{ m{3.4cm}|>{\centering}m{0.8cm}|>{\centering}m{0.8cm}|>{\centering}m{0.8cm}|>{\centering}m{0.8cm}|>{\centering}m{0.8cm}|>{\centering}m{0.8cm}|>{\centering}m{1.4cm}|>{\centering}m{1.2cm}|c}
\hline
Backbone & $AP^{b}$  & $AP^{b}_{50}$ & $AP^{b}_{75}$ & $AP^{m}$  & $AP^{m}_{50}$ & $AP^{m}_{75}$ & Params (M) & FLOPs & FPS (img/s) \\ \hline
ResNet-18\cite{he2016deep} &  34.0 & 54.0 & 36.7 & 31.2 & 51.0 & 32.7 & 31.2 & 207.4G & 42.2     \\ 
PVT-T\cite{wang2022pvt} & 36.7 & 59.2 & 39.3 & 35.1 & 56.7 & 37.3 & 32.8 & 239.8G & 26.1    \\ 
PoolFormer-S12\cite{yu2022metaformer} & 37.3 & 59.0 & 40.1 & 34.6 & 55.8 & 36.9 & 31.6 & 207.3G & 21.5 \\
\rowcolor{gray!30}
\textbf{ParFormer-T} & 37.3 & 59.0 & 40.4 & 35.2 & 56.5 & 37.5 & 25.7 & 186.7G & 44.7   \\ \hline

ResNet-50\cite{he2016deep} &  34.0 & 54.0 & 36.7 & 31.2 & 51.0 & 32.7 & 31.2 & 207.4G      & 27.9 \\
PVT-S\cite{wang2022pvt} & 40.4 & 62.9 & 43.8 & 37.8 & 60.1 & 40.3 & 44.1 &  304.5G  & 16.5   \\ 
PoolFormer-S24\cite{yu2022metaformer} & 40.1 & 62.2 & 43.4 & 37.0 & 59.1 & 39.6 & 41.0 & 239.7G & 12.7 \\
\rowcolor{gray!30}
\textbf{ParFormer-M} & 40.7 & 63.3  & 44.2 & 37.6 & 60.2 & 40.1 & 42.2 & 250.2G & 27.6  \\ \hline
\end{tabular}
\label{tab:obj}
\end{table*}

\subsubsection{SCAPE} 
The SCAPE module, which combines SCAM and Patch Embedding (PE), was assessed in three configurations to evaluate its impact on performance optimization: SCAM Before PE, SCAM Following PE and PE without SCAM module. Positioning SCAM prior to the patch embedding decreases the parameter count to 11.6 million and marginally elevates GFLOPs to 1.48 compared to the Baseline. Throughput increases marginally to 3368 images per second, while Top-1 accuracy diminishes slightly to 80.7\%. This indicates that positioning SCAM prior to PE results in inefficiencies in processing feature channels, causing a minor decrease in accuracy. Positioning SCAM subsequent to patch embedding as the baseline (11.9M and 1.48) yields a throughput of 3362 images per second. The Top-1 accuracy is maintained at 80.9\%, signifying that this configuration successfully elaborates channel-wise attention and information preservation. The placement of SCAM subsequent to PE guarantees the retention of essential features during downsampling, effectively preserving both global and local information. Eliminating SCAM entirely while preserving patch embedding reduces the parameter count to 11.5 million and enhances throughput to 3390 images per second. Nevertheless, Top-1 accuracy declines to 80.4\%, underscoring SCAM's critical function in reducing information loss across feature channels during downsampling. The lack of SCAM diminishes the model's capacity to capture and prioritize essential channel information, leading to diminished accuracy.

\subsubsection{Parallel Mixers} 
The upcoming ablation procedure is the parallel token mixer. The Parallel Mixer (PM) ratio defines how feature channels are processed by two different spatial mixing operations: depthwise convolution (DWConv) and Single Head Attention (SHA). A ratio of 0 means that all channels are processed via DWConv, while a ratio of 1/4 implies that one-quarter of the channels are handled through SHA, with the remaining 3/4 processed by DWConv. The purpose of the Parallel Mixer is to combine these two spatial mixing operations to reduce feature redundancy by leveraging DWConv for local feature extraction and SHA for global information aggregation.

The ablation study evaluates three configurations of the PM ratio in four stage architecture. The first configuration is $[0, 0, 0, 0]$. In this configuration, all channels are processed entirely by DWConv across all stages, increasing the parameter count slightly to 12.0M and reducing GFLOPs to 1.45. Throughput improves to 3435 img/s, but Top-1 accuracy drops to 80.4\%. The reliance solely on DWConv improves computational efficiency but sacrifices accuracy due to the absence of global attention provided by SHA, leading to less effective feature interaction across the entire image.
The second is [0, 0, 1/4, 1/4] that introducing SHA for 1/4 of the channels in the last two stages balances local and global feature processing. The parameter count remains at 11.9M with a slight increase in GFLOPs to 1.48. Throughput is consistent with the baseline at 3362 img/s, while Top-1 accuracy is maintained at 80.9\%. This configuration demonstrates that using SHA on a small portion of the channels in later stages helps improve global context capture without sacrificing efficiency or accuracy, making it a well-balanced approach. Finally, increasing the SHA allocation to 1/2 of the channels in the final two stages [0, 0, 1/2, 1/2] reduces the parameter count to 11.3M and GFLOPs to 1.41, while throughput improves to 3428 img/s. However, Top-1 accuracy drops to 80.2\%, indicating that while higher PM ratios increase computational efficiency and reduce model complexity, excessive reliance on SHA for global attention at the expense of DWConv’s local processing reduces the model’s ability to capture fine-grained details, leading to diminished performance.

\subsection{Object Detection and Instance Segmentation}
\subsubsection{Preparation} We assess the efficacy of ParFormer on downstream computer vision tasks. We utilize the Mask R-CNN \cite{he2017mask} algorithm to train our method on the COCO2017 train split. Subsequently, we assess the performance of the models on the val split. The training schedule is set at 1× in order to maintain a consistent comparison with earlier approaches \cite{he2016deep, chen2022mixformer, wang2022pvt, shen2021efficient}. For the 1× schedule, we conduct training for 12 epochs using a single size on the 3$\times$A6000 GPUs. AdamW\cite{kingma2014adam} Optimizer is used with an initial learning rate of $6\times10^{-5}$ with batch size 16.The training images are scaled to have a shorter side of 800 pixels and a longer side that does not exceed 1,333 pixels. The implementation is derived from the mmdetection \cite{chen2019mmdetection} code-base.  As part of the testing process, the dimension of the images is adjusted to a size of $1280\times800$ pixels. 

\begin{figure}
\centering
         \includegraphics[width=3.5cm, height=3cm]{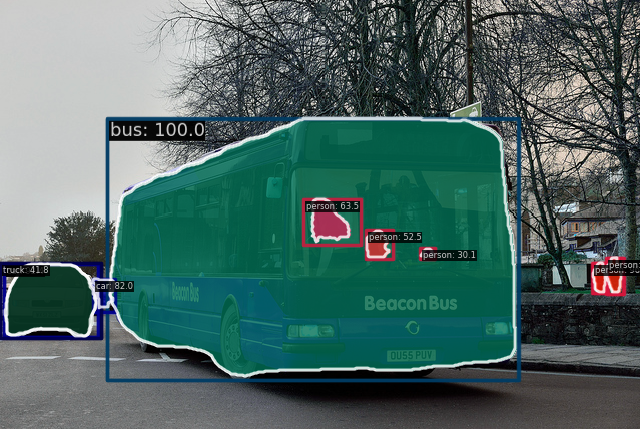}
         \includegraphics[width=3.5cm, height=3cm]{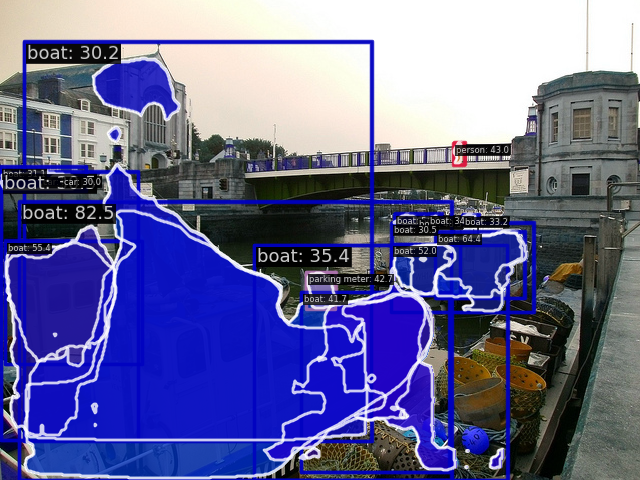}
         \includegraphics[width=3.5cm, height=3cm]{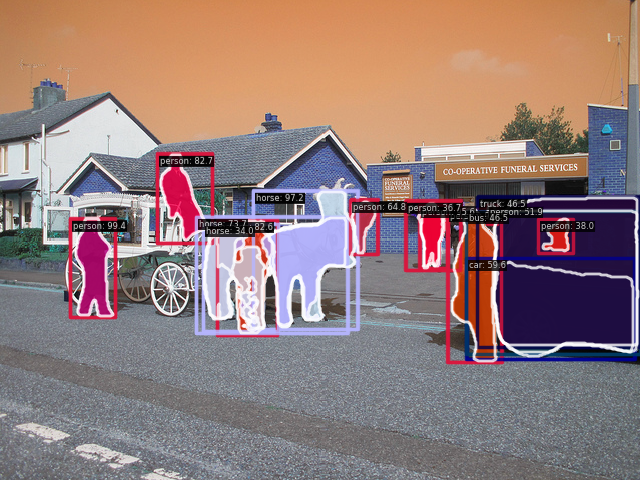}
         \includegraphics[width=3.5cm, height=3cm]{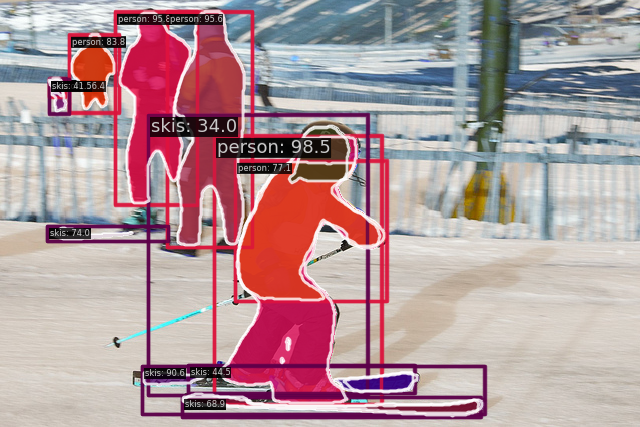}
         \includegraphics[width=3.5cm, height=3cm]{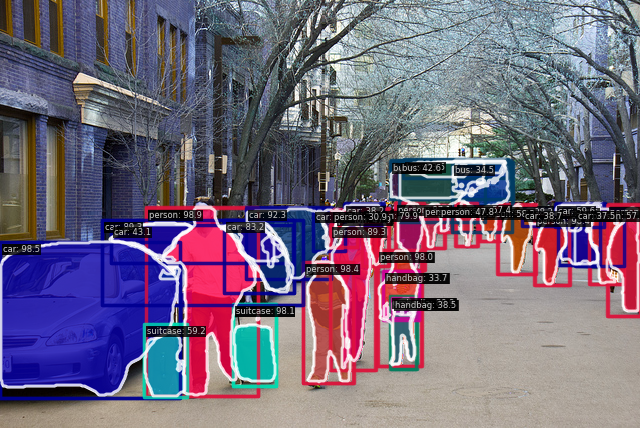}
         \includegraphics[width=3.5cm, height=3cm]{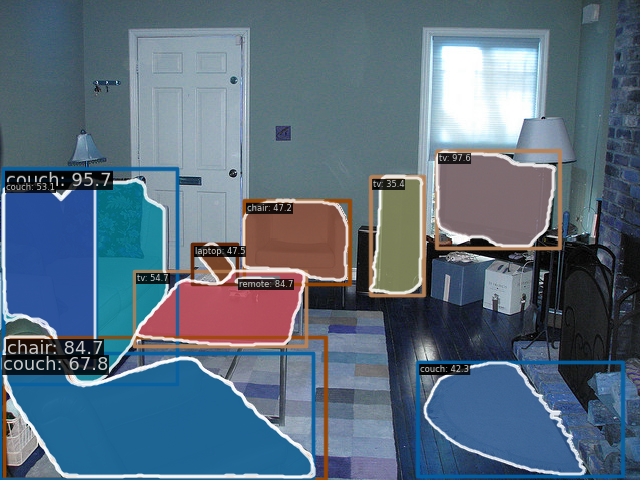}
\caption{Qualitative results of object detection and instance segmentation on COCO test2017\cite{lin2014microsoft}}
\label{fig:qualitativeobj}
\end{figure}

\subsubsection{Result} Due to computation resource constraints, we have chosen to test the ParFormer-T model for smallest model and ParFormer-M as the larger model size. Table \ref{tab:obj} compares the object detection and instance segmentation performance of various backbone networks including ParFormer on the COCO val2017 dataset using Mask R-CNN. The results are presented in terms of bounding box average precision ($AP^b$) and mask average precision ($AP^m$) at different IoU thresholds, alongside model parameters, computational complexity (FLOPs), and inference speed (FPS). The ParFormer models, particularly ParFormer-T and ParFormer-M, demonstrate superior performance across most metrics while maintaining competitive computational efficiency. Figure \ref{fig:qualitativeobj} shows the qualitative result of ParFormer as the backbone of Mask-RCNN detector for object detection and instance segmentation task in COCO dataset.

In terms of object detection, ParFormer-T achieves an ($AP^b$) of 37.3, matching the performance of ResNet-50 and PoolFormer-S12, while significantly outperforming lighter models such as ResNet-18 (34.0). ParFormer-M leads in bounding box detection, with an impressive ($AP^b$) of 40.7, surpassing other models like PVT-T (36.7), PoolFormer-S24 (40.1), and even ResNet-50 (37.3). This trend holds for higher IoU thresholds ($AP^b_{75}$), where ParFormer-M achieves the highest score of 40.1, again outperforming all other backbones.

For instance segmentation, ParFormer-T delivers strong results with an ($AP^m$) of 35.6, outperforming the baseline models such as ResNet-18 (31.2) and approaching the performance of larger backbones like PoolFormer-S12 (34.1) and ResNet-50 (35.6). ParFormer-M further excels with an ($AP^m$)of 37.6, leading over both PVT-S (37.0) and ResNet-50 (35.6), showing the efficacy of the ParFormer backbone in capturing fine-grained mask details. Notably, at the more challenging ($AP^m_{75}$) metric, ParFormer-M achieves a standout score of 40.1, the highest in the table, showcasing its ability to handle complex segmentation tasks with high precision.

\section{Conclusion}
In this paper, we introduced ParFormer, a novel vision transformer architecture designed to address the computational challenges faced by deep learning models, particularly in resource-constrained environments. By integrating a Parallel Mixer and a Sparse Channel Attention Patch Embedding (SCAPE) module, ParFormer effectively balances local and global feature extraction while reducing computational redundancy. Our approach outperforms many state-of-the-art models in terms of both accuracy and efficiency, making it highly suitable for edge-based AI applications.

Experimental results on the ImageNet-1K dataset demonstrate that ParFormer-T achieves a competitive Top-1 accuracy of 78.9\% with superior throughput performance, especially on edge devices, where it significantly outpaces competitors like MobileViT and EdgeNeXt. Additionally, ParFormer-L offers a higher Top-1 accuracy of 83.5\%, showing its scalability and suitability for high-performance tasks. On COCO object detection and segmentation, ParFormer-M achieves 40.7 AP and 37.6 AP, respectively, surpassing models like PVT-S, ResNet-50, and PoolFormer-S24 while maintaining higher efficiency.

The Parallel Mixer and SCAPE module, which are the main contributions of this work, show how powerful it can be to combine attention-based and convolutional mechanisms to get around the problems with current models. The outcomes demonstrate that ParFormer is not only a high-performance architecture but also a highly effective solution for real-time vision tasks on edge devices, which are resource- and computational-constrained. While ParFormer has proven to be an efficient and scalable model, future work could explore further optimizations, such as dynamic pruning techniques to reduce computational costs during inference. Additionally, applying ParFormer to other tasks like video classification and natural language processing could reveal its broader applicability. Lastly, investigating how this architecture can be extended to even smaller, ultra-efficient models would further enhance its utility in edge AI applications.

\ifCLASSOPTIONcaptionsoff
  \newpage
\fi

\bibliographystyle{IEEEtran}
\bibliography{egbib}

\end{document}